\documentclass{article}

\usepackage[preprint]{neurips_2023}
\usepackage{preamble}

\usepackage[utf8]{inputenc} % allow utf-8 input
\usepackage[T1]{fontenc}    % use 8-bit T1 fonts
\usepackage{hyperref}       % hyperlinks
\usepackage{url}            % simple URL typesetting
\usepackage{booktabs}       % professional-quality tables
\usepackage{amsfonts}       % blackboard math symbols
\usepackage{nicefrac}       % compact symbols for 1/2, etc.
\usepackage{microtype}      % microtypography
\usepackage{xcolor}         % colors
\usepackage{amsmath}
\usepackage{mathtools}
\usepackage{algorithm}
\usepackage{algorithmic}
\usepackage{array}
\usepackage{textcomp, mathrsfs, stmaryrd, setspace}
\usepackage{amssymb}
\usepackage{soul}
\usepackage[english]{babel}
\usepackage{subfig}
\usepackage{blindtext}
\usepackage{hyperref}
\usepackage[toc,page]{appendix}

\newtheorem{theorem}{Theorem}

\newtheorem{corollary}[theorem]{Corollary}

\title{Federated Variational Inference: Towards Improved Personalization and Generalization}
\author{
   Elahe Vedadi\thanks{Work done during an internship at Google Research.} \\
   University of Illinois at Chicago \\
   \texttt{evedad2@uic.edu} \\
   \And
   Joshua V. Dillon \\
   Google Research \\
   \texttt{jvdillon@google.com} \\
   \AND
   Philip Andrew Mansfield \\
   Google Research \\
   \texttt{memes@google.com} \\
   \And
   Karan Singhal \\
   Google Research \\
   \texttt{karansinghal@google.com } \\
   \And
   Arash Afkanpour \\
   Google Research \\
   \texttt{arashaf@google.com} \\
   \And
   Warren Richard Morningstar \\
   Google Research \\
   \texttt{wmorning@google.com} \\
}

\begin{document}

\maketitle

\newcommand{\defeq}{\stackrel{\tiny\mathrm{def}}{=}}
\newcommand{\I}{\operatorname{\mathsf I}}
\newcommand{\K}{\operatorname{\mathsf K}}
\newcommand{\E}{\operatorname{\mathsf E}}
\newcommand{\for}{\operatorname{for}}
\newcommand{\T}{\mathcal{T}}
\newcommand{\B}{\mathcal{B}}
\newcommand{\X}{\mathcal{X}}
\newcommand{\Y}{\mathcal{Y}}
\newcommand{\Z}{\mathcal{Z}}

\newcommand{\softmax}{\operatorname{softmax}}
\newcommand{\reals}{\mathbb{R}}
\newcommand{\MVN}{\operatorname{MVN}}
\newcommand{\MatrixMVN}{\operatorname{MatrixMVN}}
\newcommand{\eye}{\mathbb{I}}
\newcommand{\relu}{\operatorname{relu}}
\newcommand{\diag}{\operatorname{diag}}
\newcommand{\of}{\circ}
\newcommand{\conv}{\operatorname{conv}}
\newcommand{\dense}{\operatorname{dense}}
\newcommand{\reshape}{\operatorname{reshape}}

\begin{abstract}
%Federated learning is a privacy enhancing method that allows multiple clients to train a machine learning algorithm collaboratively under the coordination of a central server without sharing their raw data with each other. 
Conventional federated learning algorithms train a single global model by leveraging all participating clients' data. However, due to heterogeneity in client generative distributions and predictive models, these approaches may not appropriately approximate the predictive process, converge to an optimal state, or generalize to new clients. We study personalization and generalization in stateless cross-device federated learning setups assuming heterogeneity in client data distributions and predictive models. We first propose a hierarchical generative model and formalize it using Bayesian Inference. We then approximate this process using Variational Inference to train our model efficiently. We call this algorithm \emph{Federated Variational Inference (FedVI)}. We use PAC-Bayes analysis to provide generalization bounds for FedVI. We evaluate our model on FEMNIST and CIFAR-100 image classification and show that FedVI beats the state-of-the-art on both tasks.

\end{abstract}

%\textcolor{blue}{Good Key words: Algorithms, Distributed Systems, Federated Learning, Personalization}\\

%\textcolor{red}{Bad key words: Bayesian, Variational Inference,..}
%\vspace{-10pt}
\section{Introduction}
Federated Learning (FL) \citep{FederatedLearning} enables training machine learning models on decentralized datasets when privacy or other factors make it undesirable to aggregate data on a central server.  In a FL setup, the central server manages a global model that is sent to participating clients which locally train the model on their data.  After local training, model updates are aggregated and used to update the global state of the model before repeating the process.

Under simple but often contrived settings, FL can approximate centralized training and enjoy all of the same theoretical guarantees \citep[e.g., FedSGD][]{FederatedLearning}.  More realistic or practical cross-device FL setups \citep{FedOpt,wang2021field} rarely exemplify these idealized conditions.  For example, many FL algorithms perform multiple steps of local training between aggregation to minimize communication overhead. Clients in most practical FL setups participate unevenly in training, with some contributing more data than others (and many not participating at all).  Finally, each client has a separate process by which their data was generated, leading to non-independently and identically distributed (IID) client datasets.  The last issue in particular breaks theoretical guarantees of convergence, leads to a performance gap between participating and unparticipating clients \citep{yuan2022what}, and presents a challenge for training performant models in practical FL settings.

Modern approaches to address this challenge include either adding new terms to the local loss which induce convergence toward a global solution \citep{FedProx} or utilizing personalized models which address local distributional shift \citep{pFedBayes}.  Approaches for personalization have often focused on stateful FL setups, where clients are revisited throughout training and thus can update a locally stored model \citep{SCAFFOLD,wang2021field}.  However, many production scenarios are effectively stateless, since individual clients only rarely contribute to training, and local models may be either stale or non-existent.  To date, comparatively few works on personalization have focused on this setting.  Those that have \citep{FedRecon}, require clients to possess labeled examples for personalization.

In this paper, we study personalization in stateless cross-device FL setups.  We present a novel algorithm, FedVI which utilizes Variational Inference (VI) \citep{blei2017variational} to learn a model which is able to both generalize and personalize in heterogeneous client data settings. We also show that our method is able to personalize even to new clients which did not participate in training. 
We include theoretical analysis which shows that our framework bounds the Bayesian Evidence as well as the Predictive Risk.  Finally, we show experimentally that our method achieves state-of-the-art performance on two common FL benchmarks.

\textbf{Main Contributions:}
%\vspace{-5pt}
\begin{itemize}
\item We propose a hierarchical generative model based on mixed effects models for cross-device federated setups.
\item We introduce a novel algorithm, Federated Variational Inference (FedVI), to train our model efficiently in this setting.
\item  We provide generalization bounds for our proposed model using Probably Approximately Correct (PAC)-Bayes analysis.
%\item  We propose a novel neural network architecture compatible to our theoretical model, implement it in TensorFlow Federated (TFF)\footnote{\url{https://medium.com/tensorflow/introducing-tensorflow-federated-a4147aa20041}} framework, and scale up the implementation to NVIDIA Tesla V100 GPUs for hyper parameter tuning.
\item  We evaluate the FedVI algorithm on two federated datasets, FEMNIST and CIFAR-100, and show that FedVI beats the previous state-of-the-art methods on both datasets.
\end{itemize}

%Federated learning \citep[FL;][]{FederatedLearning} is an emerging approach that enables different clients to train a machine learning model using all clients’ data under coordination of a central server, without the need to transfer each client’s data to the central server. One important challenge in this area is the heterogeneity of clients’ data distributions (\cite{SCAFFOLD, TighterTheoryLocalSGD}). In particular, due to the data distributions heterogeneity, training to minimize the local objective at each client, may not lead to minimizing the global objective after aggregation of all clients’ updates.  In this paper our goal is to train machine learning models using variational Bayesian inference in a cross- device federated setup to improve generalization of the trained models on a finite distributed population of clients, and personalization of the trained models for new clients. 

\section{Related Work}\label{sec:relatedworks}

\textbf{Bayesian FL:}  In order to address some of the problems of statistical heterogeneity in FL, several works used Bayesian approaches to encode additional domain knowledge or help induce convergence.  Early attempts \citep{PPFL, FedBE} focused on model aggregation, either to retain uncertainty in model parameters, or to weight parameter updates proportional to performance.  \cite{pFedBayes} instead attempt to use a Bayesian Neural Network (BNN) approximated with VI to train a global model using a Kullback–Leibler (KL) regularizer which induces convergence similar to the proximal term in FedProx \citep{FedProx}. While their local models can, in principle, personalize by deviating from the global model, they realistically require stateful settings with significant labeled data on clients in order to do so.  \cite{FedPop} casts personalized FL as mixed effects regression, and attempts to model the inherent heterogeneity in this setting explicitly using Stochastic Gradient Langevin Dynamics \citep{SGLD}.  Our proposed method assumes a similar generative process to \cite{FedPop} but instead uses VI to efficiently infer the posterior, as well as place a bound on the predictive risk to induce generalization to new clients \citep{PAC-Bayes-Germain}.

\textbf{Stateful FL:} There is a rich body of literature on personalization in FL  \citep{clusteredFL, FedAvgplus, FedRep, APFL, ditto}. Many previous approaches focus on stateful settings, where a set of local parameters is stored on clients and is maintained throughout rounds of training.  In contrast, we focus on stateless settings where it is not possible to maintain an up-to-date local state on each client.  This is similar to the setting considered by \cite{KNN-Per}, who use K-nearest neighbors to account for client distributional shift.  While this is a robust means of dealing with both input and output distributional shift, it requires clients to possess labeled examples for every class (which is unrealistic in real-world setups), and cannot be used outside of classification problems.

\textbf{Meta Learning:} %  
There is a significant amount of prior work that studies connections between personalized FL and Model-Agnostic Meta-Learning (MAML) approaches \citep{MAML, FedRecon, per-fedavg, FedRep, lin2023meta, chen2019federated}. The main idea behind these works is to find an initial global shared model that the existing or new clients can adapt to their own dataset by performing a few steps of gradient descent with respect to their local data. FedRecon \citep{FedRecon} is also motivated by MAML and considers a partially local federated learning setting, where only a subset of model parameters (known as global parameters) will be aggregated and trained globally for fast reconstruction of the local parameters. Our work can be considered as an extension of FedRecon.  Unlike this work, we also provide a means of reconstructing local parameters without access to labeled data.

%\vspace{-10pt}
\section{Methods}
\vspace{-5pt}
\subsection{Hierarchical Generative Model}
\vspace{-5pt}
Let us consider a stateless cross-device federated setup with a population of multiple clients and a central server where at each round a randomly selected subset of clients participate in training. In this setup we first group the model parameters of each client into global and local parameters; $\theta$ and $\beta_k$, for $k \in [c]$,\footnote{In this paper we represent the set of $\{1,\dots,c\}$ by $[c]$.} respectively, where $c$ denotes the total number of clients in the system. Global parameters are aggregated and updated globally at the server at the end of each round of training, while local parameters never leave the clients. We assume that global parameters are a single sample from the prior distribution $t(\Theta)$, and that each client’s local parameters are an independent sample from the local prior $r(B_k)$.
Moreover, we assume that the data may not be IID between clients \emph{i.e}, $x_{ik} \sim\nu_k(X_k)$ for $i  \in [n_k]$ and $k \in [c]$, where $n_k$ represents the total number of data samples at client $k$. We also assume that each client can have a different predictive distribution \emph{i.e}, while all clients have the same family of likelihood distribution (predictive model) $\ell(Y|f(\theta, \beta_k, x_{ik}))$, the distribution depends on $\beta_k$ and thus can be different for each client.

The above setup is a prototypical example of a mixed effects model \citep{demidenko2013mixed}, which is often used in scenarios where we want to predict one continuous random variable based on multiple independent (random or fixed) factors using repeated measures from the same unit of observation.  Mixed effects models \citep{demidenko2013mixed} are extremely well studied, so by formalizing our setup in this context we can take advantages of the theoretical underpinnings we already have in this space.  

To summarize, we propose the following hierarchical data generating process:
% Mixed effects models \citep{demidenko2013mixed} are well studied and often used in scenarios where we want to predict one continuous random variable based on multiple independent (random or fixed) variables while we have repeated measures from the same unit of observation. Therefore, one can see that our considered federated setup has the required criteria to be formalized by mixed effects models. Inspired by that, we propose the following hierarchical data generating process:
\begin{align} \label{eq:hierarchical-model}
&\theta \sim t(\Theta)\\ \nonumber 
&\for k \in[c]:\\ \nonumber 
&\quad \beta_k \sim r(B_k)\\ \nonumber 
&\quad \for i\in[n_k]:\\ \nonumber
&\quad \quad x_{ik} \sim\nu_k(X_k)\\ \nonumber 
&\quad\quad y_{ik}\sim \ell(Y|f(\theta, \beta_k, x_{ik})),
\end{align}
where %$c$ is the number of clients at each round of training, $n_k$ is the number of data samples stored at client $k$ (for $k \in [c]$\footnote{We note that in this paper we represent the set of $\{1,\dots,c\}$ by $[c]$.}), and 
$f$:$ \Phi \times \B_k \times \X_k \to \Z$ is a deterministic function (e.g., DNN) mapping what we know to the parameters of $\ell(.)$, our distribution over outcomes.

\begin{figure*}
		\centering
		\includegraphics[width=11cm]{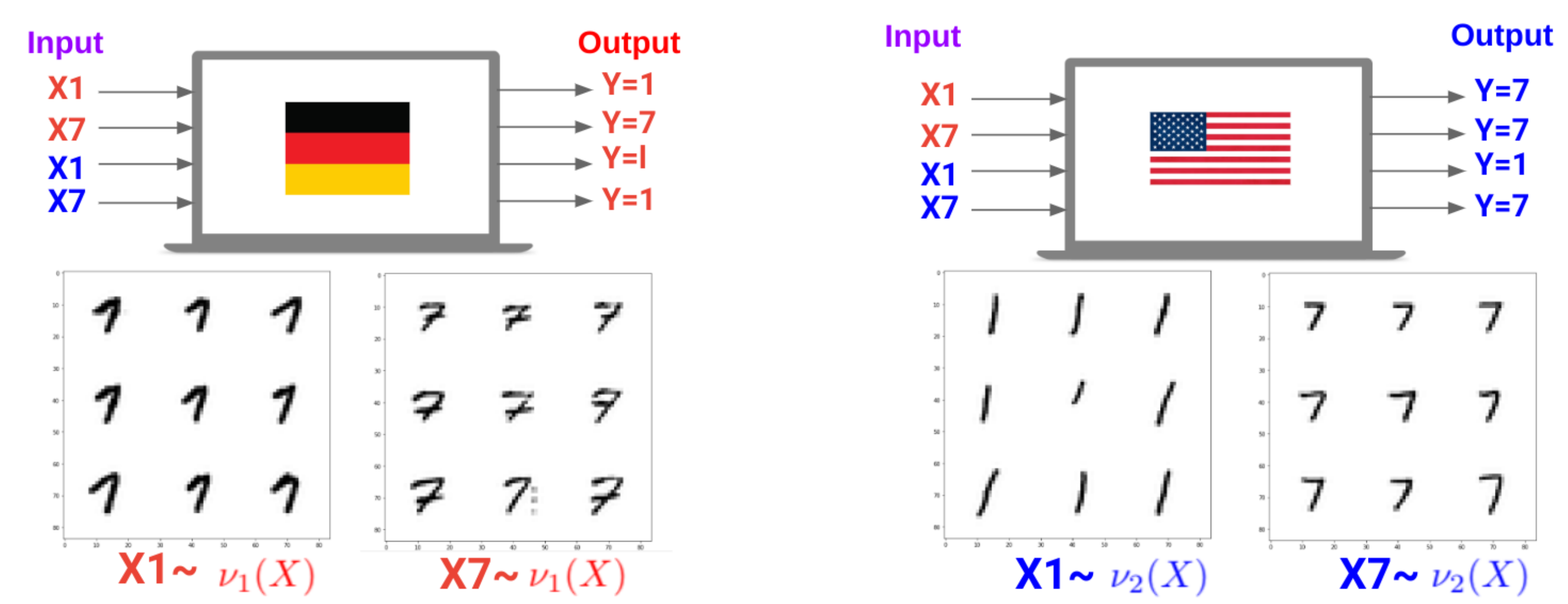}
	\caption{Intuitive illustration of different data generating distributions and predictive models in cross-device federated setups. 
	}
\label{fig:German-vs-American-1&7}
\vspace{-10pt}
\end{figure*}

To intuitively understand the effect of differing data generating processes and predictive distributions, consider the Federated EMNIST dataset (FEMNIST; Figure \ref{fig:German-vs-American-1&7}), where each client's dataset is composed of numbers and letters written by that client.  The input data for each client incorporates their particular writing style; for example a German client may write their sevens with a horizontal middle bar, while an American client may not.  Similarly, the German client may place a hood on top of a $1$ while the American client may not.  This describes the difference in data generating distributions.  This also illustrates that each client may have different predictive distributions: the American client may see the German's $1$ as a $7$, while the German client may see the American's $1$ as a lowercase "l".  Thus their predictive distributions are in direct conflict with each other.  An entirely global model cannot account for this variation and thus needs to have some degree of local modification in order to properly model the data generating process.

Our proposed algorithm explicitly assumes this data generating process.  Note that this assumption reduces in special cases to existing FL setups, such as IID predictive distributions ($r(B_k)=\delta(B_k - \beta)$), or IID data generating proceses ($\nu_k(X_k) = \nu(X_k)$).  In the following section, we detail how we use VI to efficiently infer the model parameters.
%\vspace{-10pt}
\subsection{Training Procedure}
\vspace{-5pt}
% Given the proposed process above and choices for the model architecture, likelihood, and prior, one has everything they need to know in order to define the posterior, and by extension the posterior predictive distribution $\hat{p}(Y)\defeq p(Y|\{ x^{n_k}\}^c)$.  In practice, defining either of these are rarely straightforward.  This section is thus dedicated to defining a simple and efficient procedure for approximating the posterior predictive distribution and using it to make predictions on new data.
The main goal of this section is to clarify how we train the proposed hierarchical model to be able to make predictions. In other words for making predictions we need to calculate the estimated probability density function of labels (given input data), $\hat{p}(Y)\defeq p(Y|\{ x^{n_k}\}^c)$, which is defined as the following \citep{watanabe2018mathematical}:
\begin{align}
    \hat{p}(Y) \defeq \int_{\theta} \int_{\beta_c} \dots \int_{\beta_1} p(\theta, B^c|\{y^{n_k}, x^{n_k}\}^c) \ell(Y|f(\theta, \{\beta_k, x^{n_k}\}^c)),
\end{align}
where $B^c \defeq \{\beta_k\}^c \defeq \{\beta_k: k \in [c]\}$, $x^{n_k} \defeq \{x_i: i\in [n_k]\}$, and $\{x^{n_k}\}^c \defeq \{x_{ik}: i\in[n_k], k\in [c]\}$. Therefore, for calculating $\hat{p}(Y)$ it is required to calculate the posterior probability of model parameters given the training data.  Based on the fact that for arbitrary random variables $x$, $y$, and $z$ we have $p(x,y|z) = p(x|y,z)p(y|z)$, one can rewrite the posterior probability of the model parameters given training data as follows:
\begin{align}\label{eq:quantity-of-interest}
p(\theta,B^c|\{y^{n_k},x^{n_k}\}^c)
&=\frac{p(\theta,B^c,\{y^{n_k}\}^c|\{x^{n_k}\}^c)}{p(\{y^{n_k}\}^c|\{x^{n_k}\}^c)}.
\end{align}

Assuming that the prior distribution of the global parameters, $t(\theta)$, the prior distribution of the local parameters, $r(\beta_k)$, and the likelihood distribution of each client, $\ell(Y|f(\theta, \beta_k, x_{ik}))$, are independent we calculate the joint probability distribution as follows.
\begin{align}
p(\theta,B^c,\{y^{n_k}\}^c|\{x^{n_k}\}^c)
&= p(\theta,\{\beta_k,\{y_{ik}\}_{i\in[n_k]}\}_{k\in[c]}|\{x_{ik}\}_{k\in[c],i\in[n_k]}) \nonumber \\
&= t(\theta)\prod_{k\in[c]}r(\beta_k)\prod_{i\in[n_k]}\ell(y_{ik}|f(\theta,\beta_k,x_{ik})) \nonumber \\
&= t(\theta)\prod_{k\in[c]}r(\beta_k)\prod_{k\in[c]}\prod_{i\in[n_k]}\ell(y_{ik}|f(\theta,\beta_k,x_{ik})) \nonumber \\
&= t(\theta)r(B^c)\ell(Y|f(\theta,B^c,X)),
\end{align}
where $X = \{x^{n_k}\}^c$ and $Y = \{y^{n_k}\}^c$ . Considering the fact that for arbitrary random variables $x$, $y$, and $z$ we have $p(y|z) = \int_{x} p(x,y|z)$, the marginal distribution can be written as the following.
\begin{align}
p(\{y^{n_k}\}^c|\{x^{n_k}\}^c) = \int_{\theta} \int_{\beta_c} \dots \int_{\beta_1} p(\theta,B^c,\{y^{n_k}\}^c|\{x^{n_k}\}^c).
\end{align}

Unfortunately this integral is not only infeasible to compute, but also mathematically intractable, \emph{i.e}, there is no closed form solution. Consequently, this intractable denominator makes the whole fraction intractable. 
One way to solve this problem is to approximate this intractable posterior, $p(\theta,B^c|\{y^{n_k},x^{n_k}\}^c)$, with a tractable surrogate distribution, $q(\theta,B^c|\{y^{n_k},x^{n_k}\}^c)$, and borrowing ideas from VI we can find the best candidate for the surrogate distribution by devising a specific upper bound on the marginal distribution, which is called ELBO \citep{AutoEncodeVB} and is equal to the KL divergence between the posterior and the surrogate distributions (Equation \ref{eq:ELBO-upper-bound}). Minimizing the ELBO upper bound with respect to the surrogate distribution, $q(\theta,B^c|\{y^{n_k},x^{n_k}\}^c)$, gives us the best approximation for the intractable posterior distribution, $p(\theta,B^c|\{y^{n_k},x^{n_k}\}^c)$. In the following equation $\mathsf{K}[.]$ stands for the KL divergence between two distributions. The detailed derivations of this equation is provided in Appendix \ref{appendix:A}. %\ref{appendix:A}.% in the supplemental materials.

\vspace{-12pt}
\begin{align}
\label{eq:ELBO-upper-bound}
-\log p(\{y^{n_k}\}^c|\{x^{n_k}\}^c)
&\le \min_q \mathsf{K}[q(\theta,B^c|\{y^{n_k},x^{n_k}\}^c),p(\theta,B^c,\{y^{n_k}\}^c|\{x^{n_k}\}^c)].
\end{align}
%\arash{$\mathsf{K}$ is not defined above.}\elahe{done}
By asserting factorization, we define the surrogate as a parametric distribution as the following:

\vspace{-15pt}
\begin{align}
q(\theta,B^c|\{y^{n_k},x^{n_k}\}^c)
&\defeq q_\lambda(\theta|\underbrace{\{y^{n_k},x^{n_k}\}^c}_{\defeq D})\prod_{k\in[c]}q_\lambda(\beta_k|\theta,\underbrace{y^{n_k},x^{n_k}}_{\defeq D_k}) %\nonumber \\
\defeq q_\lambda(\theta|D)q_\lambda(B^c|\theta,D),
\end{align}
where $\lambda$ is the parameter set that uniquely defines the surrogate distribution. Therefore, the objective function that we need to minimize to train our proposed hierarchical model is the ELBO upper bound, which can be written as follows based on the definition of KL divergence, properties of logarithms, and multiplication rule.

\vspace{-10pt}
\begin{align}\label{eq:loss}
\mathcal{J}(\lambda;\gamma, \tau)
&= \mathsf{K}[q_\lambda(\theta,B^c|\{y^{n_k},x^{n_k}\}^c),p(\theta,B^c,\{y^{n_k}\}^c|\{x^{n_k}\}^c)] \nonumber \\
&= \sum_{k\in[c]}\sum_{i\in[n_k]} \overbrace{\mathsf{E}_{q_\lambda(\theta|D)q_\lambda(\beta_k|\theta,D_k)}\left[-\log \ell(y_{ik}|f(\theta,\beta_k,x_{ik}))\right]}^\text{Per Datum Expected Loss}
 \nonumber\\&\quad
 + \underbrace{\gamma\mathsf{K}[q(\theta|D),t(\theta)]}_\text{Global Regularizer}
 %\nonumber\\&\quad
 + \sum_{k\in[c]} \tau \underbrace{\mathsf{E}_{q_\lambda(\theta|D)} \mathsf{K}[q_\lambda(\beta_k|\theta,D_k),r(\beta_k)],}_\text{Local Regularizer}
\end{align}

where $\gamma$, $\tau$, $t(\theta),r(\beta_k)$, and the functional form of $q_\lambda(\theta,B^c|D)$ are left as hyper parameters. The details of this derivation are provided in Appendix \ref{appendix:B}. %\ref{appendix:B}.% in the supplemental materials.
%\vspace{-5pt}
\section{Generalization Bounds}
%\vspace{-5pt}
As we described in the previous section, the objective function that we chose to minimize to train our proposed hierarchical model is the ELBO upper bound. By minimizing this objective function, in the best scenario (\emph{i.e}, if we could find the global minimum) we are able to minimize the error on the training dataset (empirical risk). However, for the sake of generalization guarantees our goal is to minimize the error on unseen datasets (true risk, or generalization error). For this purpose we did PAC-Bayes analysis based on  the results of Theorem 3 in \citep{PAC-Bayes-Germain} to calculate the generalization bound on the true risk of our model.
 
\begin{corollary}
\label{coro:PAC-Bayes}
Given a distribution $\nu$ over $\{X, Y\}=\{x^{n_k}, y^{n_k}\}^c$, a hypothesis set $\mathcal{F}=\{\theta, B^c\}$, a loss
function $\ell : \mathcal{F} \times \X \times \Y \to \mathbb{R}$, a prior distribution $\pi(\Theta, \B^c)=t(\Theta)r(\B^c)$ over $\mathcal{F}$, a $\delta \in (0,1]$ and a real number $\eta>0$,
with probability at least $1-\delta$ over the choice of $(X,Y) \sim \nu^c$, for any $q$ on $\mathcal{F}$ we have:
\begin{align}\label{eq:coroll-1}
    &\overbrace{\mathsf{E}_\nu [-\log\big(\mathsf{E}_{q(\theta,B^c|X,Y)}[\ell(Y|X,\theta,B^c)\big)]]}^\text{True risk} \leq \nonumber \\
    &\overbrace{\mathsf{E}_{\nu^c}[\mathsf{E}_{q(\theta,B^c|X,Y)}[-\log(\ell(Y|X,\theta,B^c))]]}^\text{Empirical risk} 
    + \frac{1}{\eta}\bigg[\overbrace{\mathsf{K}[q(\theta,B^c|X,Y), \pi(\theta, B^c)]}^\text{KL divergence} \nonumber \\
    &+\underbrace{\log\big(\tfrac{1}{\delta}\mathsf{E}_{\nu^c}\mathsf{E}_{\pi(\theta,B^c)}\big[\exp\bigg(\eta\mathsf{E}_\nu[-\log(\ell(Y|X,\theta,B^c))]-\eta\mathsf{E}_{\nu^c}[-\log(\ell(Y|X,\theta,B^c))]\bigg)\big]\big)}_\text{Slack term}\bigg].
\end{align}
\end{corollary}
 
{\textbf{Sketch of Proof:}} The proof of this corollary directly follows the proof of Theorem 3 in (\cite{PAC-Bayes-Germain}) and is derived with the help of Jensen inequality, Donsker-Varadhan change of measure inequality, and Markov's inequality. Details are provided in Appendix \ref{appendix:C}.

Now that we have the generalization bound of our model on the right side of Equation \ref{eq:coroll-1}, we can see that it is equal to the ELBO upper bound (Equation \ref{eq:loss}) plus a slack term, which is independent of the surrogate and the posterior distributions. Therefore, under the condition of having a finite slack term, minimizing the ElBO upper bound with respect to the surrogate distribution is equivalent to minimizing the generalization error with respect to the surrogate distribution. Thus, we conclude that with the assumption of having a finite slack term, and with probability greater than $1-\delta$, minimizing the ELBO upper bound ensures generalization for our model.

%\section{Information Theoretic Perspective}\label{sec:infothe}

%Information harvesting is one of the important aspects of each learning model, which tells us how much information our model needs to learn to be able to make predictions. Therefore, using the definition of mutual information, we can rewrite the ELBO upper bound, Equation \ref{eq:loss}, in terms of mutual information as the following.
%\begin{align}
%\label{eq:info_theo_loss}
 %\mathcal{J}(\lambda;\gamma,\tau)=\gamma I(\theta;D) + \tau I(B^c;\theta,D)-I(Y;\theta,B^c,X)
%\end{align}
%where $\gamma$ and $\tau$ are assumed as the hyper parameters.

%\textbf{\textcolor{blue}{Proof}:} The proof is provided in Appendix \ref{appendix:D}.

%As we can see, by minimizing this loss function we are minimizing the mutual information between model parameters and training data, and maximizing the mutual information between labels and the joint of model parameters and training data; which is a mathematical representation of what we discussed in Section \ref{sec:relatedworks-infothe}.

%\vspace{-8pt}
\section{Implementation and Experimental Evaluation}\label{sec:eval}
%The main goal of this section is to go through the details of our primary theoretical assumptions and model architecture for implementing and evaluating our proposed FedVI algorithm.
%\vspace{-5pt}
\textbf{Distributions:} For the prior distribution of the local parameters, we assume a normal distribution with zero mean and variance equal to that given by the initialization scheme \citep[e.g.][]{Glorot2010UnderstandingTD, pmlr-v15-glorot11a, He2015DelvingDI}.
We do not make any assumptions about the data generating distribution of each client. We use a categorical distribution as our likelihood, where the logits generated by a deep neural network parameterized by $\theta$ and $\beta$ (described below). To simplify implementation, we used a point estimate for the global posterior and used a uniform non-normalized distribution for the global prior (i.e. $\log t(\Theta)=0$).  This is equivalent to assuming zero weight for the KL penalty, and we found that this appears to work in our experiments but note that one may need to use a stronger prior when there is more significant variation in the input data between clients.

\begin{figure*}
		\centering
		\includegraphics[width=12cm]{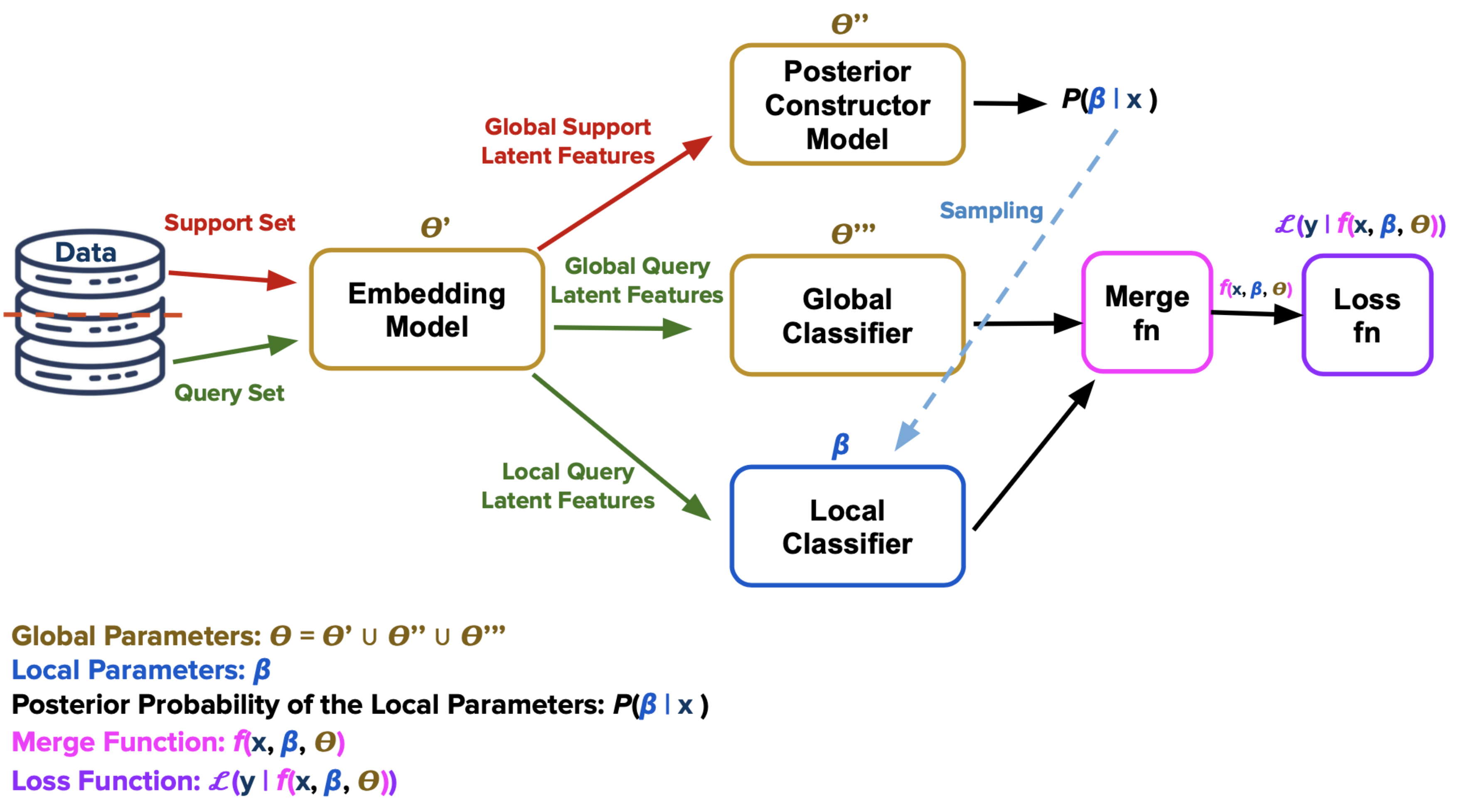}
	\caption{Our proposed model architecture implementing FedVI. 
	}
\label{fig:model_arc}
%\vspace{-10pt}
\end{figure*}

%\vspace{-10pt}
%\subsection{Datasets}
%\vspace{-5pt}
\textbf{Tasks:} We evaluate our FedVI algorithm on two different datasets, FEMNIST\footnote{\url{https://www.tensorflow.org/federated/api_docs/python/tff/simulation/datasets/emnist/load_data}} \citep{caldas2019leaf} ($62$-class digit and character classification) and CIFAR-100\footnote{\url{https://www.tensorflow.org/federated/api_docs/python/tff/simulation/datasets/cifar100/load_data}} \citep{Krizhevsky2009LearningML} ($100$-class classification). FEMNIST is particularly relevant since it has a naturally different data generative distribution for each client. Although CIFAR-100 data is synthetically partitioned using a hierarchical Latent Dirichlet Allocation (LDA) process \citep{li2006pachinko} and distributed among clients, we evaluated our algorithm on this dataset as well to show the superiority of our method on a more complicated classification task.
%\vspace{-10pt}
%\subsection{Implementation}
%\vspace{-5pt}

\textbf{Implementation:} We implement our FedVI algorithm in TensorFlow Federated (TFF) and scale up the implementation to NVIDIA Tesla V100 GPUs for hyperparameter tuning. For FEMNIST dataset with $3400$ clients we consider the first $20$ clients as non-participating users which are held-out in training to better measure generalization as in \citep{yuan2022what}. At each round of training we select $100$ clients uniformly at random without replacement, but with replacement across rounds. For CIFAR-100 with $500$ training clients, we set the data of the first $10$ clients as holdout data and select $50$ clients uniformly at randomly at each round. We train FedVI algorithm on both FEMNIST and CIFAR-100 for $1500$ rounds and at each round of training we divide both datasets into mini-batches of $256$ data samples and used mini-batch gradient descent algorithm to optimize the objective function.  
%\vspace{-10pt}
%\subsection{Architecture, Optimizers, and Hyper Parameters}\label{subsec:arc-opt-hp}
% \vspace{-5pt}

\textbf{Model Architecture:} There are infinitely many model architectures which could implement our method. The architecture we used in our experiments is illustrated in Figure \ref{fig:model_arc}. It consists of four separate modules: an embedding model which encodes the input as a vector, a posterior reconstruction model which predicts the posterior over local parameters, a classifier parameterized by global parameters (implemented as a single dense layer), and a classifier implemented by local parameters (generated by sampling from the reconstructed posterior).  A forward pass through this model is as follows:
\begin{enumerate}
\item Input data is fed into the embedding model to extract vector representations of the data.
\item The representations are partitioned over the batch dimension into support and query sets.  Similar to FedRecon \citep{FedRecon}, the support set is used to reconstruct the local parameters and the query set is used to make predictions.  Note that the support set we use can be unlabeled, and that the two sets need not be disjoint.  However, we used disjoint sets in our experiments since \citep{FedRecon} found that it improved their model performance.
\item The representation for both support and query sets are further split over their features axis into local and global features.
\item The global features of the support set are used to reconstruct the local posterior. The local parameters are generated by sampling from this posterior.
\item The global features of the query set are passed to the global classifier to get the global predictions, and the local features of the query set and local parameters are passed to the local classifier to get the local modifications to the global prediction.
\item The local and global predictions are merged to get the predictions.  The log-likelihood is then computed between these predictions and the label and added to the KL divergence between local posterior and prior.
\end{enumerate}

\textbf{Data Partitioning:} First we note that for both FEMNIST and CIFAR-100 datasets, at each epoch we consider the first $50\%$ of each mini-batch as the support set and the other $50\%$ as the query set (\emph{i.e}, for a mini-batch with $256$ data samples the first $128$ samples belong to the support set and the rest belong to query set). For the global-local features split, we found that using a larger number of global features ($80\%$) than local features ($20\%$) performed best.

\textbf{Embedding Model:} For the embedding model for FEMNIST, to facilitate comparisons to previous works, we used the network architecture from \citep{FedOpt}. To replicate their global model exactly, we only used their first 2 layers of their architecture as our embedding model, while the final layer of their model was used as our global classifier. For CIFAR-100 we choose a similar model but with $5$ convolution layers with channels $(32, 64, 128, 256, 512)$.

\textbf{Posterior Constructor Model, Global, and Local Classifiers:} The posterior constructor model, global, and local classifiers are the same for both FEMNIST and CIFAR-100 experiments. We use a $3$ layer MLP parameterized by the global parameters for the posterior constructor model, and one dense layer for each of global and local classifiers parameterized by global and local parameters, respectively. 

\begin{figure}[tp] 
    \centering
   % \subfloat[data a]{%
   \vspace{-10pt} 
        \includegraphics[width=0.6\textwidth]{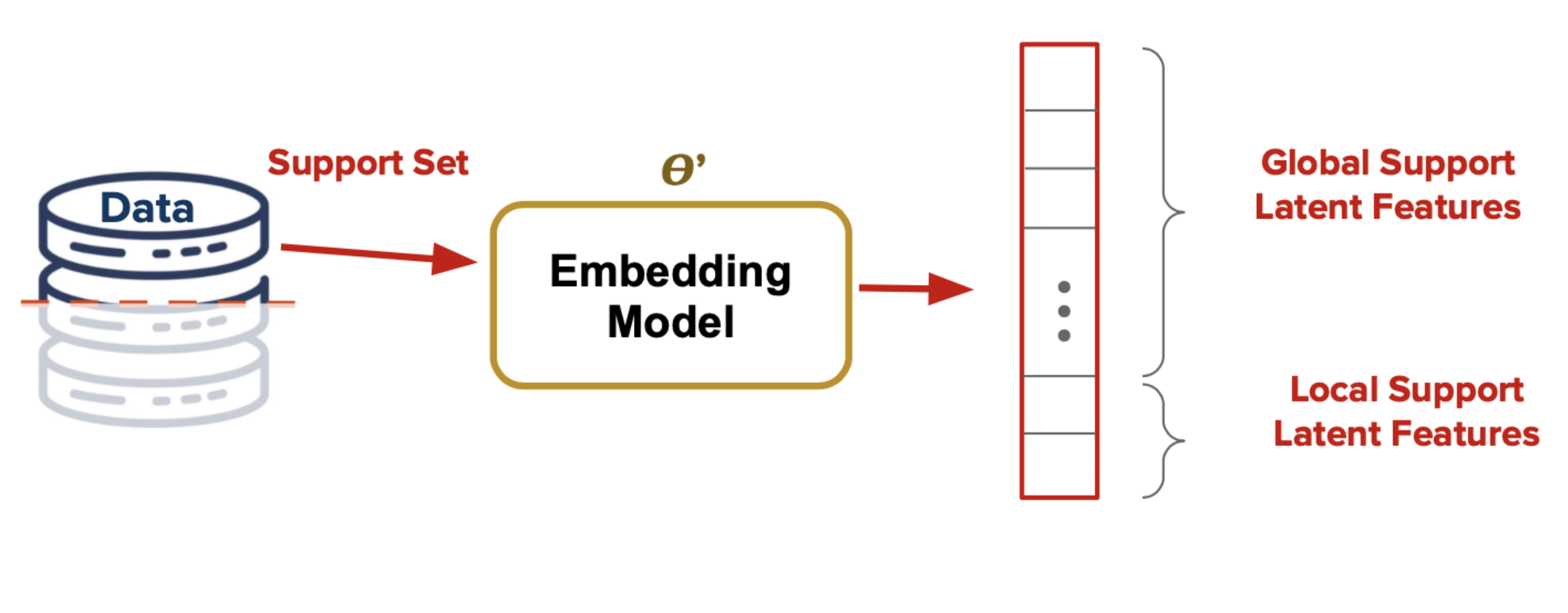}%
    %
    %    }%
  %  \hfill%
   % \subfloat[data b]{%
     %   \includegraphics[width=0.7\textwidth]{Figures/Querydata.png}%
      %  \label{fig:b}%
      %  }%
    \vspace{-15pt}  
    \caption{An illustration of the division of the data into support and query sets, as well as the division into global and local features.  Note that the actual split into support and query sets happens after the forward pass of the embedding model.}
    \label{fig:divide_glob_loc_features}
\end{figure}

\textbf{Optimizers:} We use Stochastic Gradient Descent (SGD) for our client optimizer and SGD with momentum for the server optimizer for all experiments \citep{FedOpt}. We set the client learning rate equal to $0.03$ for CIFAR-100 and $0.02$ for FEMNIST dataset, and server learning rate equal to $3.0$ with momentum $0.9$ for both FEMNIST and CIFAR-100 datasets. 

Additional details about the empirical evaluation are provided in Appendix \ref{appendix:D}.

\vspace{-9pt}
\subsection{Evaluation Results and Discussion}
\vspace{-5pt}
We compare our proposed FedVI algorithm with a state-of-the-art personalized FL method, KNN-Per \citep{KNN-Per}, as well as other methods including FedAvg \citep{FederatedLearning}, FedAvg+ \citep{FedAvgplus}, ClusteredFL \citep{clusteredFL}, DITTO \citep{ditto}, FedRep \citep{FedRep}, and APFL \citep{APFL}, using the results reported in \citep{KNN-Per}. 
\begin{figure}[htp]
    \centering
   % \subfloat[data a]{%
        \includegraphics[width=0.6\textwidth]{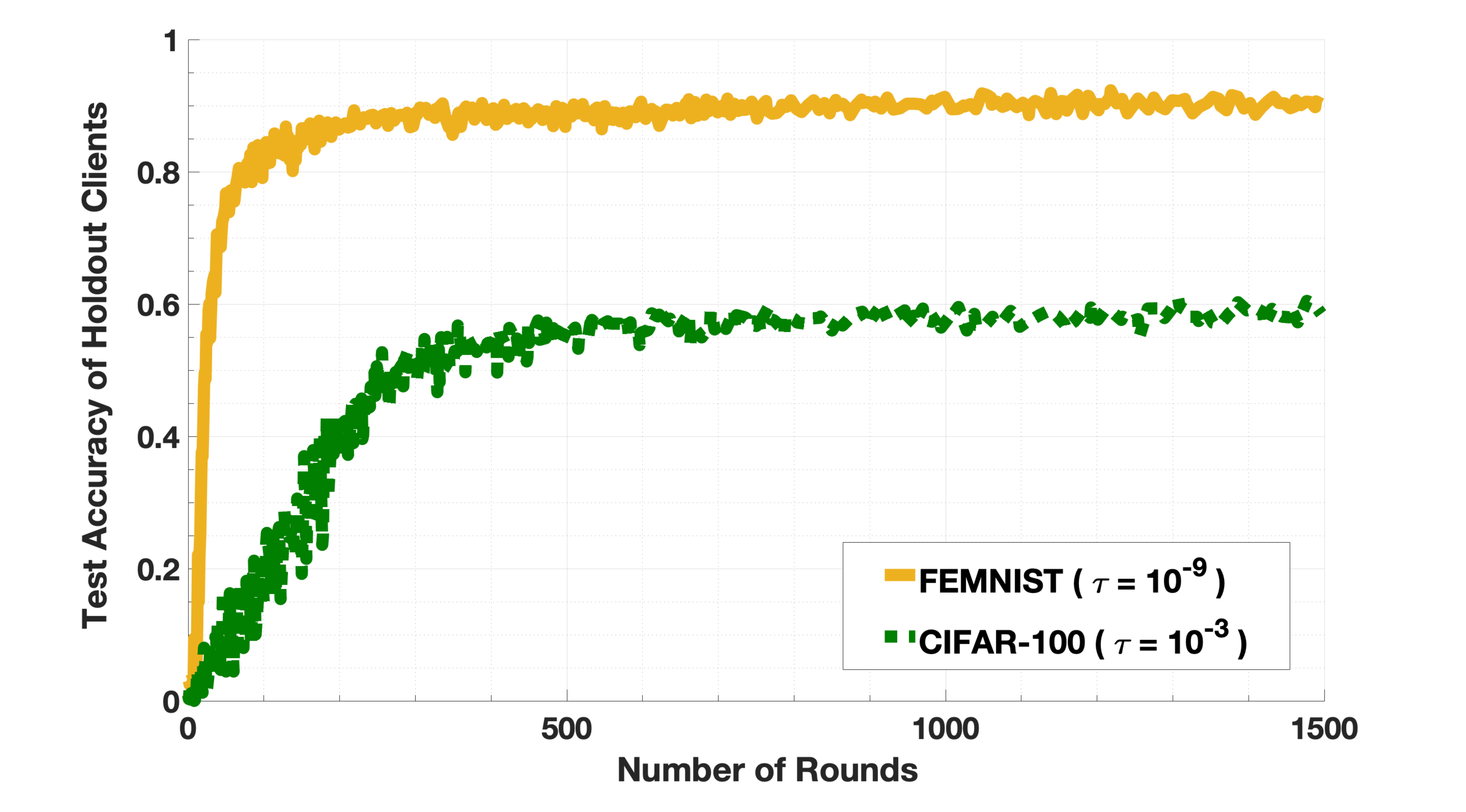}%
    %
    %    }%
  %  \hfill%
   % \subfloat[data b]{%
     %   \includegraphics[width=0.7\textwidth]{Figures/Querydata.png}%
      %  \label{fig:b}%
      %  }%
    \caption{Non-participating test accuracy of FEMNIST and CIFAR-100 for $1500$ rounds of training.}
    \label{fig:unpart-test-acc}
\end{figure}

The performance our proposed FedVI algorithm and other methods on the local test dataset of each client (unseen data at training) are provided in Tables \ref{tabl:test-acc} and \ref{tabl:Non-part-test-acc} for participating and non-participating (completely unseen during training) clients, respectively. All of the reported values are average weighted accuracy with weights proportional to local dataset sizes. To ensure the robustness of our reported results for FedVI, we average test accuracy across the last $100$ rounds of training. We also show non-participating test accuracy over 1500 rounds of training in Figure \ref{fig:unpart-test-acc}. FedVI achieves the state-of-the-art participating and non-participating test accuracy on both FEMNIST and CIFAR-100 datasets.

\begin{table}
  \caption{Test accuracy of the clients who participate in training.}
  \label{tabl:test-acc}
  \centering
  \resizebox{\textwidth}{!}{\begin{tabular}{lllllllll}
    %\toprule
    %\multicolumn{2}{c}{Part}                   \\
    \cmidrule(r){1-9}
    Dataset     & FedAvg & FedAvg+ & ClusteredFL & DITTO & FedRep & APFL & KNN-Per & FedVI \\
    \midrule
    FEMNIST & 83.4   & 84.3 & 83.7 & 84.3 & 85.3 & 84.1 & 88.2 & \textbf{90.3}     \\
    CIFAR-100 & 47.4 & 51.4 & 47.2 & 52.0 & 53.2 & 51.7 & 55.0 & \textbf{59.1}   \\
    \bottomrule
  \end{tabular}}
\end{table}
%\vspace{-20pt}
\begin{table}
  \caption{Test accuracy of the non-participating clients.}
  \label{tabl:Non-part-test-acc}
  \centering
  \resizebox{\textwidth}{!}{\begin{tabular}{lllllllll}
    %\toprule
    %\multicolumn{2}{c}{Part}                   \\
    \cmidrule(r){1-9}
    Dataset     & FedAvg & FedAvg+ & ClusteredFL & DITTO & FedRep & APFL & KNN-Per & FedVI \\
    \midrule
    FEMNIST & 83.1 & 84.2 & 83.2 & 83.9 & 85.4 & 84.2 & 88.1 & \textbf{90.6}     \\
    CIFAR-100 & 47.1 & 50.8 & 47.1 & 52.1 & 53.5 & 49.1 & 56.1 & \textbf{58.7}   \\
    \bottomrule
  \end{tabular}}
\end{table}

\begin{figure}[htp] 
    %\centering
    \vspace{-15pt}
    \subfloat[FEMNIST]{%
    \includegraphics[width=0.51\textwidth]{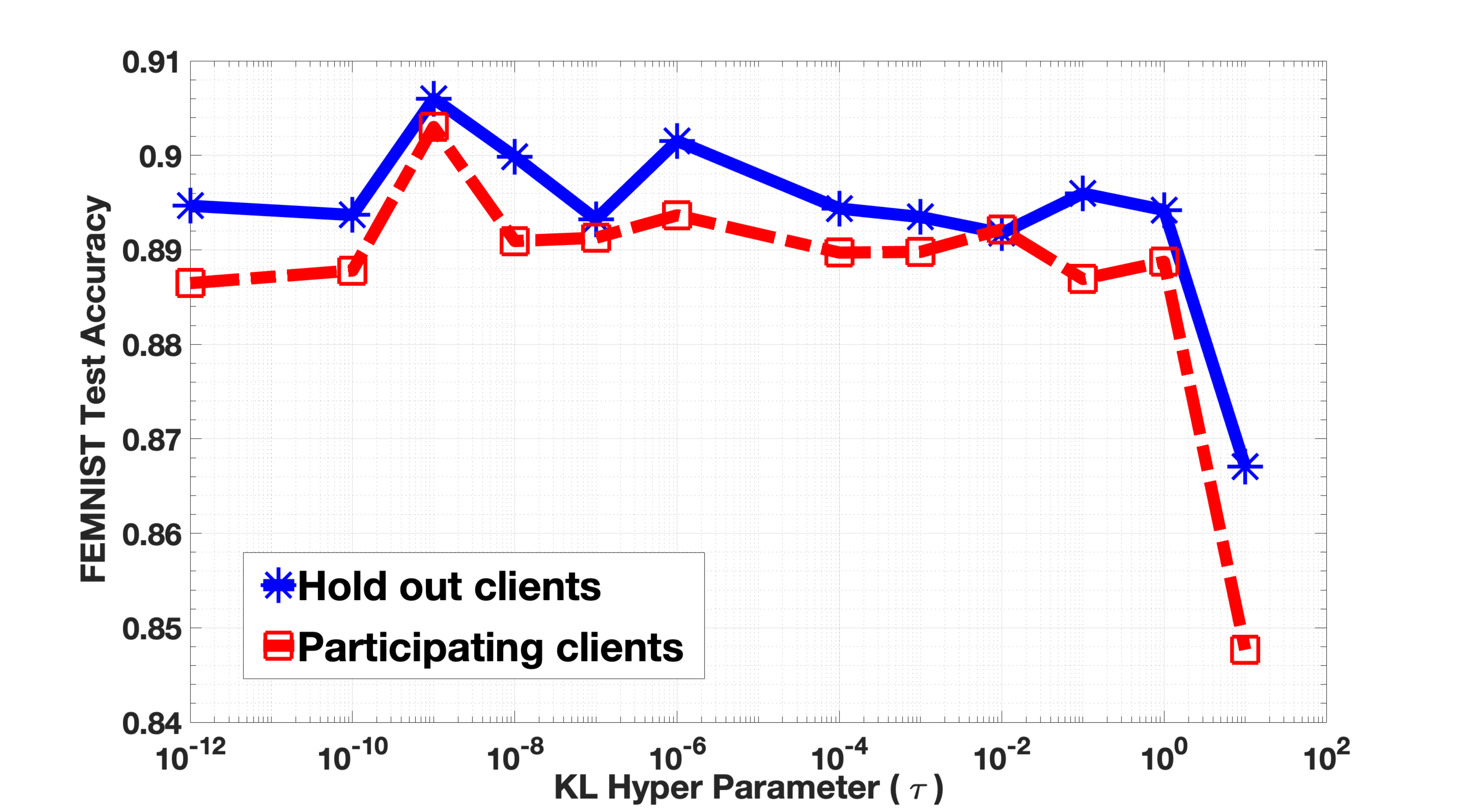}%
        \label{fig:femnist-acc-vs-KLhp}
        }%[width=14cm] or [width=0.5\textwidth]
    \hfill%
    \subfloat[CIFAR-100]{%
     \includegraphics[width=0.51\textwidth]{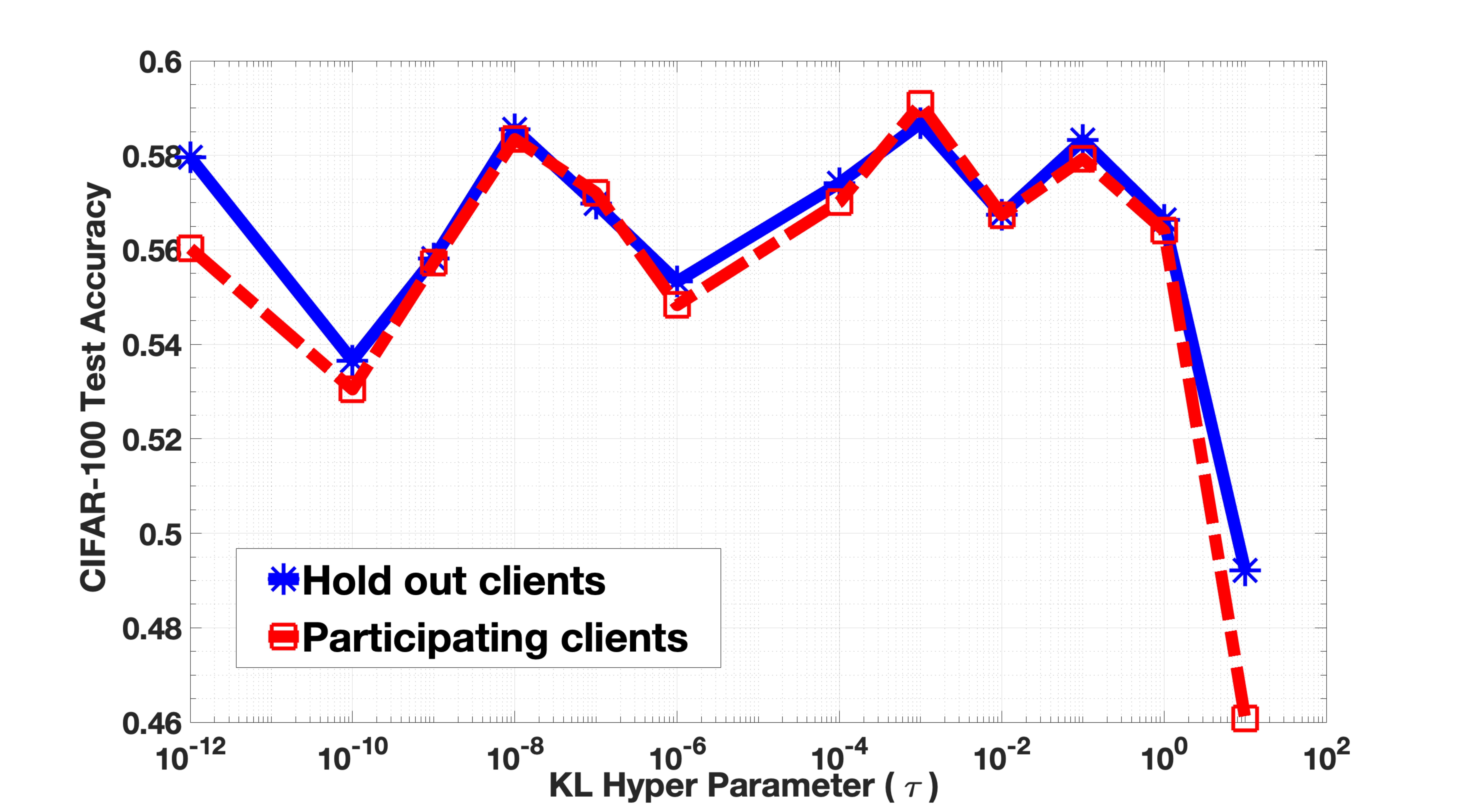}%
        \label{fig:cifar-acc-vs-KLhp}
        }%
    \caption{Participating and non-participating test accuracy vs. KL hyperparameter $\tau$. }
    \label{fig:acc-vs-KLhp}
\end{figure}

Figure \ref{fig:femnist-acc-vs-KLhp} is an ablation for KL hyperparameter $\tau$. We plot average test accuracy over the last $100$ rounds of FEMNIST training vs. KL hyperparameter in range of $\tau \in \{0, 10^{-10}, 10^{-9}, \dots, 0.1, 1, 10\}$\footnote{As the horizontal axis of both figures in Figure \ref{fig:acc-vs-KLhp} are semi-logarithmic, test accuracy results of $\tau=0$ are shown at point $\tau=10^{-12}$.}. We see that $\tau=10^{-9}$ is the best hyperparameter choice for FEMNIST dataset and gives us much higher accuracy with a smaller generalization gap compared to $\tau=0$.

Figure \ref{fig:cifar-acc-vs-KLhp} illustrates average of the last $100$ rounds of test accuracy vs. KL hyperparameter $\tau$ for CIFAR-100. $\tau=10^{-3}$ has the highest participating and non-participating accuracy. Moreover by comparing the results of $\tau=0$ with other values ($\tau \neq 0$), we can see that minimizing KL divergence decreases the participation test accuracy gap, as expected. Comparing this figure with Figure \ref{fig:femnist-acc-vs-KLhp}, we see that the difference between test accuracy of $\tau=0$ and $\tau=10^{-9}$ in FEMNIST experiment is notably greater than the difference between test accuracy of $\tau=0$ and $\tau=10^{-3}$ in CIFAR-100 experiment. This means that minimizing KL divergence is more important for FEMNIST compared to CIFAR-100. A potential reason is that for FEMNIST, the data generating distribution of each client is naturally different, while for CIFAR-100 we synthetically partition data and distribute it among clients.
%\vspace{-10pt}
\vspace{-10pt}
%(\textcolor{blue}{how many bits of info FedVI learns??information theoretic discussion??}) %\begin{figure*}[htp] 
 %   \begin{minipage}[t]{0.61\linewidth}
  %  \includegraphics[width=\linewidth]{Figures/cifar100-kl.png}
   % \end{minipage}\hfill
    %\begin{minipage}[t]{0.61\linewidth}
    %\includegraphics[width=\linewidth]{Figures/femnist-kl.png}
    %\end{minipage}
%\end{figure*}

\section{Conclusion and Future Work}
\vspace{-5pt}
In this work we focus on the problem of personalization in stateless cross-device federated setups. We proposed a novel algorithm, FedVI, which is based on mixed effects models and is trained via Variational Inference. Then we provide generalization bounds for FedVI based on PAC-Bayes analysis. Afterwards, we introduce a novel architecture for FedVI and implement it. Finally, we evaluate FedVI on FEMNIST and CIFAR-100, and show that it beats the state-of-the-art on both datasets. 

In this paper we learn a point estimate of the global parameters instead of their prior distribution. In future work, we plan to explore other priors for the global parameters to provide more formal guarantees of generalization. We also aim to use non-disjoint support and query sets to see how this affects the performance of Fed-VI algorithm. Regarding our proposed model architecture in Section \ref{sec:eval}, as we mentioned before, it is one of many possible architectures that is compatible with our theoretical hierarchical model. Further work can refine the model architecture to achieve improved results. To investigate our proposed algorithm from an information theoretic point of view, we provide an information theoretic representation of our objective function (Equation \ref{eq:loss}) in Appendix \ref{appendix:E}. Future work could also explore connections to differential privacy to be able to provide privacy guarantees for our model.

\bibliographystyle{plainnat}
\newpage
\bibliography{references}

\begin{thebibliography}{38}
\providecommand{\natexlab}[1]{#1}
\providecommand{\url}[1]{\texttt{#1}}
\expandafter\ifx\csname urlstyle\endcsname\relax
  \providecommand{\doi}[1]{doi: #1}\else
  \providecommand{\doi}{doi: \begingroup \urlstyle{rm}\Url}\fi

\bibitem[Alemi(2020)]{pmlr-v118-alemi20a}
Alexander~A. Alemi.
\newblock Variational predictive information bottleneck.
\newblock In Cheng Zhang, Francisco Ruiz, Thang Bui, Adji~Bousso Dieng, and
  Dawen Liang, editors, \emph{Proceedings of The 2nd Symposium on Advances in
  Approximate Bayesian Inference}, volume 118 of \emph{Proceedings of Machine
  Learning Research}, pages 1--6. PMLR, 08 Dec 2020.
\newblock URL \url{https://proceedings.mlr.press/v118/alemi20a.html}.

\bibitem[Alemi and Fischer(2018)]{ThermoDynamML}
Alexander~A. Alemi and Ian Fischer.
\newblock Therml: Thermodynamics of machine learning, 2018.
\newblock URL \url{https://arxiv.org/abs/1807.04162}.

\bibitem[Alemi et~al.(2017)Alemi, Fischer, Dillon, and Murphy]{alemi2017deep}
Alexander~A. Alemi, Ian Fischer, Joshua~V. Dillon, and Kevin Murphy.
\newblock Deep variational information bottleneck.
\newblock In \emph{International Conference on Learning Representations}, 2017.
\newblock URL \url{https://openreview.net/forum?id=HyxQzBceg}.

\bibitem[Blei et~al.(2017)Blei, Kucukelbir, and McAuliffe]{blei2017variational}
David~M Blei, Alp Kucukelbir, and Jon~D McAuliffe.
\newblock Variational inference: A review for statisticians.
\newblock \emph{Journal of the American statistical Association}, 112\penalty0
  (518):\penalty0 859--877, 2017.

\bibitem[Caldas et~al.(2019)Caldas, Duddu, Wu, Li, Konečný, McMahan, Smith,
  and Talwalkar]{caldas2019leaf}
Sebastian Caldas, Sai Meher~Karthik Duddu, Peter Wu, Tian Li, Jakub Konečný,
  H.~Brendan McMahan, Virginia Smith, and Ameet Talwalkar.
\newblock Leaf: A benchmark for federated settings, 2019.

\bibitem[Chen et~al.(2019)Chen, Luo, Dong, Li, and He]{chen2019federated}
Fei Chen, Mi~Luo, Zhenhua Dong, Zhenguo Li, and Xiuqiang He.
\newblock Federated meta-learning with fast convergence and efficient
  communication, 2019.

\bibitem[Chen and Chao(2020)]{FedBE}
Hong-You Chen and Wei-Lun Chao.
\newblock Fedbe: Making bayesian model ensemble applicable to federated
  learning.
\newblock \emph{arXiv preprint arXiv:2009.01974}, 2020.

\bibitem[Chen and Chao(2022)]{FedAvgplus}
Hong-You Chen and Wei-Lun Chao.
\newblock On bridging generic and personalized federated learning for image
  classification, 2022.

\bibitem[Collins et~al.(2023)Collins, Hassani, Mokhtari, and
  Shakkottai]{FedRep}
Liam Collins, Hamed Hassani, Aryan Mokhtari, and Sanjay Shakkottai.
\newblock Exploiting shared representations for personalized federated
  learning, 2023.

\bibitem[Cuff and Yu(2016)]{Cuff_2016}
Paul Cuff and Lanqing Yu.
\newblock Differential privacy as a mutual information constraint.
\newblock In \emph{Proceedings of the 2016 {ACM} {SIGSAC} Conference on
  Computer and Communications Security}. {ACM}, oct 2016.
\newblock \doi{10.1145/2976749.2978308}.
\newblock URL \url{https://doi.org/10.1145%2F2976749.2978308}.

\bibitem[Demidenko(2013)]{demidenko2013mixed}
Eugene Demidenko.
\newblock \emph{Mixed models: theory and applications with R}.
\newblock John Wiley \& Sons, 2013.

\bibitem[Deng et~al.(2020)Deng, Kamani, and Mahdavi]{APFL}
Yuyang Deng, Mohammad~Mahdi Kamani, and Mehrdad Mahdavi.
\newblock Adaptive personalized federated learning, 2020.

\bibitem[Fallah et~al.(2020)Fallah, Mokhtari, and Ozdaglar]{per-fedavg}
Alireza Fallah, Aryan Mokhtari, and Asuman~E. Ozdaglar.
\newblock Personalized federated learning: {A} meta-learning approach.
\newblock \emph{CoRR}, abs/2002.07948, 2020.
\newblock URL \url{https://arxiv.org/abs/2002.07948}.

\bibitem[Finn et~al.(2017)Finn, Abbeel, and Levine]{MAML}
Chelsea Finn, Pieter Abbeel, and Sergey Levine.
\newblock Model-agnostic meta-learning for fast adaptation of deep networks.
\newblock \emph{CoRR}, abs/1703.03400, 2017.
\newblock URL \url{http://arxiv.org/abs/1703.03400}.

\bibitem[Germain et~al.(2016)Germain, Bach, Lacoste, and
  Lacoste-Julien]{PAC-Bayes-Germain}
Pascal Germain, Francis Bach, Alexandre Lacoste, and Simon Lacoste-Julien.
\newblock Pac-bayesian theory meets bayesian inference.
\newblock In D.~Lee, M.~Sugiyama, U.~Luxburg, I.~Guyon, and R.~Garnett,
  editors, \emph{Advances in Neural Information Processing Systems}, volume~29.
  Curran Associates, Inc., 2016.
\newblock URL
  \url{https://proceedings.neurips.cc/paper/2016/file/84d2004bf28a2095230e8e14993d398d-Paper.pdf}.

\bibitem[Ghosh et~al.(2021)Ghosh, Chung, Yin, and Ramchandran]{clusteredFL}
Avishek Ghosh, Jichan Chung, Dong Yin, and Kannan Ramchandran.
\newblock An efficient framework for clustered federated learning, 2021.

\bibitem[Glorot and Bengio(2010)]{Glorot2010UnderstandingTD}
Xavier Glorot and Yoshua Bengio.
\newblock Understanding the difficulty of training deep feedforward neural
  networks.
\newblock In \emph{International Conference on Artificial Intelligence and
  Statistics}, 2010.

\bibitem[Glorot et~al.(2011)Glorot, Bordes, and Bengio]{pmlr-v15-glorot11a}
Xavier Glorot, Antoine Bordes, and Yoshua Bengio.
\newblock Deep sparse rectifier neural networks.
\newblock In Geoffrey Gordon, David Dunson, and Miroslav Dudík, editors,
  \emph{Proceedings of the Fourteenth International Conference on Artificial
  Intelligence and Statistics}, volume~15 of \emph{Proceedings of Machine
  Learning Research}, pages 315--323, Fort Lauderdale, FL, USA, 11--13 Apr
  2011. PMLR.
\newblock URL \url{https://proceedings.mlr.press/v15/glorot11a.html}.

\bibitem[He et~al.(2015)He, Zhang, Ren, and Sun]{He2015DelvingDI}
Kaiming He, X.~Zhang, Shaoqing Ren, and Jian Sun.
\newblock Delving deep into rectifiers: Surpassing human-level performance on
  imagenet classification.
\newblock \emph{2015 IEEE International Conference on Computer Vision (ICCV)},
  pages 1026--1034, 2015.

\bibitem[Karimireddy et~al.(2019)Karimireddy, Kale, Mohri, Reddi, Stich, and
  Suresh]{SCAFFOLD}
Sai~Praneeth Karimireddy, Satyen Kale, Mehryar Mohri, Sashank~J. Reddi,
  Sebastian~U. Stich, and Ananda~Theertha Suresh.
\newblock Scaffold: Stochastic controlled averaging for federated learning,
  2019.
\newblock URL \url{https://arxiv.org/abs/1910.06378}.

\bibitem[Kingma and Welling(2013)]{AutoEncodeVB}
Diederik~P Kingma and Max Welling.
\newblock Auto-encoding variational bayes, 2013.
\newblock URL \url{https://arxiv.org/abs/1312.6114}.

\bibitem[Kotelevskii et~al.(2022)Kotelevskii, Vono, Moulines, and
  Durmus]{FedPop}
Nikita Kotelevskii, Maxime Vono, Eric Moulines, and Alain Durmus.
\newblock Fedpop: A bayesian approach for personalised federated learning,
  2022.
\newblock URL \url{https://arxiv.org/abs/2206.03611}.

\bibitem[Krizhevsky(2009)]{Krizhevsky2009LearningML}
Alex Krizhevsky.
\newblock Learning multiple layers of features from tiny images.
\newblock 2009.

\bibitem[Li et~al.(2020)Li, Sahu, Zaheer, Sanjabi, Talwalkar, and
  Smith]{FedProx}
Tian Li, Anit~Kumar Sahu, Manzil Zaheer, Maziar Sanjabi, Ameet Talwalkar, and
  Virginia Smith.
\newblock Federated optimization in heterogeneous networks, 2020.

\bibitem[Li et~al.(2021)Li, Hu, Beirami, and Smith]{ditto}
Tian Li, Shengyuan Hu, Ahmad Beirami, and Virginia Smith.
\newblock Ditto: Fair and robust federated learning through personalization,
  2021.

\bibitem[Li and McCallum(2006)]{li2006pachinko}
Wei Li and Andrew McCallum.
\newblock Pachinko allocation: Dag-structured mixture models of topic
  correlations.
\newblock In \emph{Proceedings of the 23rd international conference on Machine
  learning}, pages 577--584, 2006.

\bibitem[Lin et~al.(2023)Lin, Ren, Chen, Ren, Yu, Ma, de~Rijke, and
  Cheng]{lin2023meta}
Yujie Lin, Pengjie Ren, Zhumin Chen, Zhaochun Ren, Dongxiao Yu, Jun Ma, Maarten
  de~Rijke, and Xiuzhen Cheng.
\newblock Meta matrix factorization for federated rating predictions, 2023.

\bibitem[Marfoq et~al.(2022)Marfoq, Neglia, Kameni, and Vidal]{KNN-Per}
Othmane Marfoq, Giovanni Neglia, Laetitia Kameni, and Richard Vidal.
\newblock Personalized federated learning through local memorization, 2022.

\bibitem[McMahan et~al.(2016)McMahan, Moore, Ramage, Hampson, and
  Arcas]{FederatedLearning}
H.~Brendan McMahan, Eider Moore, Daniel Ramage, Seth Hampson, and Blaise
  Agüera~y Arcas.
\newblock Communication-efficient learning of deep networks from decentralized
  data.
\newblock 2016.
\newblock \doi{10.48550/ARXIV.1602.05629}.
\newblock URL \url{https://arxiv.org/abs/1602.05629}.

\bibitem[Reddi et~al.(2020)Reddi, Charles, Zaheer, Garrett, Rush, Konečný,
  Kumar, and McMahan]{FedOpt}
Sashank Reddi, Zachary Charles, Manzil Zaheer, Zachary Garrett, Keith Rush,
  Jakub Konečný, Sanjiv Kumar, and H.~Brendan McMahan.
\newblock Adaptive federated optimization, 2020.
\newblock URL \url{https://arxiv.org/abs/2003.00295}.

\bibitem[Singhal et~al.(2021)Singhal, Sidahmed, Garrett, Wu, Rush, and
  Prakash]{FedRecon}
Karan Singhal, Hakim Sidahmed, Zachary Garrett, Shanshan Wu, Keith Rush, and
  Sushant Prakash.
\newblock Federated reconstruction: Partially local federated learning, 2021.
\newblock URL \url{https://arxiv.org/abs/2102.03448}.

\bibitem[Thorgeirsson and Gauterin(2020)]{PPFL}
Adam~Thor Thorgeirsson and Frank Gauterin.
\newblock Probabilistic predictions with federated learning.
\newblock \emph{Entropy}, 23\penalty0 (1):\penalty0 41, 2020.

\bibitem[Tishby et~al.(2000)Tishby, Pereira, and Bialek]{tishby2000information}
Naftali Tishby, Fernando~C. Pereira, and William Bialek.
\newblock The information bottleneck method, 2000.

\bibitem[Wang et~al.(2021)Wang, Charles, Xu, Joshi, McMahan, Al-Shedivat,
  Andrew, Avestimehr, Daly, Data, et~al.]{wang2021field}
Jianyu Wang, Zachary Charles, Zheng Xu, Gauri Joshi, H~Brendan McMahan, Maruan
  Al-Shedivat, Galen Andrew, Salman Avestimehr, Katharine Daly, Deepesh Data,
  et~al.
\newblock A field guide to federated optimization.
\newblock \emph{arXiv preprint arXiv:2107.06917}, 2021.

\bibitem[Watanabe(2018)]{watanabe2018mathematical}
Sumio Watanabe.
\newblock \emph{Mathematical theory of Bayesian statistics}.
\newblock CRC Press, 2018.

\bibitem[Welling and Teh(2011)]{SGLD}
Max Welling and Yee~W Teh.
\newblock Bayesian learning via stochastic gradient langevin dynamics.
\newblock In \emph{Proceedings of the 28th international conference on machine
  learning (ICML-11)}, pages 681--688, 2011.

\bibitem[Yuan et~al.(2022)Yuan, Morningstar, Ning, and Singhal]{yuan2022what}
Honglin Yuan, Warren~Richard Morningstar, Lin Ning, and Karan Singhal.
\newblock What do we mean by generalization in federated learning?
\newblock In \emph{International Conference on Learning Representations}, 2022.
\newblock URL \url{https://openreview.net/forum?id=VimqQq-i_Q}.

\bibitem[Zhang et~al.(2022)Zhang, Li, Li, Guo, and Shao]{pFedBayes}
Xu~Zhang, Yinchuan Li, Wenpeng Li, Kaiyang Guo, and Yunfeng Shao.
\newblock Personalized federated learning via variational bayesian inference,
  2022.

\end{thebibliography}
\newpage
\begin{appendices}

\section{Derivations of Equation \ref{eq:ELBO-upper-bound}}
\label{appendix:A}
Here we provide the detailed derivations of Equation \ref{eq:ELBO-upper-bound} which are derived based on Section 2.2 of \citep{AutoEncodeVB}. The main goal of these derivations is to devise an upper bound on the negative logarithm of the intractable denominator of Equation \ref{eq:quantity-of-interest} to be able to approximate the posterior probability of the model parameters $p(\theta,B^c|\{y^{n_k},x^{n_k}\}^c)$, in a tractable way. For this purpose, we consider an arbitrary distribution $q(\theta,B^c|\{y^{n_k},x^{n_k}\}^c)$ as a surrogate for the posterior. Because the KL divergence of two distributions is always non-negative, we can use the KL divergence between the true posterior and our surrogate to devise an obvious and trivial upper bound on $-\log p(\{y^{n_k}\}^c|\{x^{n_k}\}^c)$ as the initial step in Equation \ref{subeq:appA-1}. As the minimum of a non-negative number is always non-negative, we replace the KL divergence with its minimum value with respect to the surrogate distribution $q(\theta,B^c|\{y^{n_k},x^{n_k}\}^c)$, to make this upper bound as tight as possible (Equation \ref{subeq:appA-2}). Moreover, since $-\log p(\{y^{n_k}\}^c|\{x^{n_k}\}^c)$ is independent of the surrogate distribution, we move this term inside the minimum as shown in Equation \ref{subeq:appA-3}. The rest of the proof comes from the definition of KL divergence, the multiplication rule of probability, and properties of logarithms. For the sake of simplicity in notation we have $\{y^{n_k},x^{n_k}\}^c \defeq D$ in the following equations.
\begin{align}
-\log p(\{y^{n_k}\}^c|\{x^{n_k}\}^c)
&\le -\log p(\{y^{n_k}\}^c|\{x^{n_k}\}^c) %\nonumber \\
+  \overbrace{\mathsf{K}[q(\theta,B^c|D),p(\theta,B^c|D)]}^\text{Always $\ge 0$.} \label{subeq:appA-1} \\ 
\Rightarrow -\log p(\{y^{n_k}\}^c|\{x^{n_k}\}^c)
&\le -\log p(\{y^{n_k}\}^c|\{x^{n_k}\}^c)+ \overbrace{\min_q \mathsf{K}[q(\theta,B^c|D),p(\theta,B^c|D)]}^\text{Always $\ge 0$.} \label{subeq:appA-2} \\
\Rightarrow -\log p(\{y^{n_k}\}^c|\{x^{n_k}\}^c)
&\le \min_q -\log p(\{y^{n_k}\}^c|\{x^{n_k}\}^c) %\nonumber \\
+ \mathsf{K}[q(\theta,B^c|D),p(\theta,B^c|D)] \label{subeq:appA-3} \\
&= \min_q \mathsf{E}_{q(\theta,B^c|D)}[-\log p(\{y^{n_k}\}^c|\{x^{n_k}\}^c) + \log \frac{q(\theta,B^c|D)}{p(\theta,B^c|D)}] \nonumber \\
&= \min_q \mathsf{E}_{q(\theta,B^c|D)}[ \log \frac{q(\theta,B^c|D)}{p(\theta,B^c|D)p(\{y^{n_k}\}^c|\{x^{n_k}\}^c)}] \nonumber \\
&= \min_q \mathsf{E}_{q(\theta,B^c|D)}[ \log \frac{q(\theta,B^c|D)}{p(\theta,B^c,\{y^{n_k}\}^c|\{x^{n_k}\}^c)}] \nonumber \\
&=\min_q \mathsf{K}[q(\theta,B^c|D),p(\theta,B^c,\{y^{n_k}\}^c|\{x^{n_k}\}^c)].
\end{align}

\section{Derivations of Equation \ref{eq:loss}}
\label{appendix:B}

We provide details for Equation \ref{eq:loss}, which is derived based on the definition of KL divergence, properties of logarithms, and the multiplication rule of probability. In the following equations $\{y^{n_k},x^{n_k}\}^c \defeq D$ for the simplicity in notations.
\begin{align}
 p(\{y^{n_k}\}^c|\{x^{n_k}\}^c)
& =\frac{p(\theta,B^c,\{y^{n_k}\}^c|\{x^{n_k}\}^c)}{p(\theta,B^c|\{y^{n_k},x^{n_k}\}^c)} %\nonumber \\
 = \frac{p(\theta,B^c,\{y^{n_k}\}^c|\{x^{n_k}\}^c)}{p(\theta,B^c|D)} \times \frac{q(\theta,B^c|D)}{q(\theta,B^c|D)} \nonumber \\
&= \frac{p(\theta,B^c,\{y^{n_k}\}^c|\{x^{n_k}\}^c)}{q(\theta,B^c|D)} \times \frac{q(\theta,B^c|D)}{p(\theta,B^c|D)} \nonumber \\
&= \frac{t(\theta)r(B^c)\ell(Y|f(\theta,B^c,X))}{q_\lambda(\theta|D)q_\lambda(B^c|\theta,D)} \times \frac{q(\theta,B^c|D)}{p(\theta,B^c|\theta,D)} \nonumber \\
&\Rightarrow -\log(p(\{y^{n_k}\}^c|\{x^{n_k}\}^c))
= -\log(\ell(Y|f(\theta,B^c,X))) \nonumber \\
&+ \log (\frac{q_\lambda(\theta|D)}{t(\theta)})+\log(\frac{q_\lambda(B^c|\theta,D)}{r(B^c)})-\log(\frac{q(\theta,B^c|D)}{p(\theta,B^c|\theta,D)}) \nonumber \\
&\Rightarrow \mathsf{E}_{q(\theta,B^c|D)}[-\log(p(\{y^{n_k}\}^c|\{x^{n_k}\}^c))] = -\log(p(\{y^{n_k}\}^c|\{x^{n_k}\}^c)) \nonumber \\
&= \mathsf{E}_{q(\theta,B^c|D)}[-\log(\ell(Y|f(\theta,B^c,X)))]
+ \mathsf{E}_{q(\theta,B^c|D)}[\log (\frac{q_\lambda(\theta|D)}{t(\theta)})] \nonumber %\\
\end{align}
\begin{align}
&+\mathsf{E}_{q(\theta,B^c|D)}[\log(\frac{q_\lambda(B^c|\theta,D)}{r(B^c)})]-\mathsf{E}_{q(\theta,B^c|D)}[\log(\frac{q(\theta,B^c|D)}{p(\theta,B^c|D)})] \nonumber \\
& \Rightarrow -\log(p(\{y^{n_k}\}^c|\{x^{n_k}\}^c))+\mathsf{K}[q(\theta,B^c|D),p(\theta,B^c|D)] \nonumber  \\
&=  \mathsf{E}_{q(\theta,B^c|D)}[-\log(\ell(Y|f(\theta,B^c,X)))] \nonumber \\
&+\mathsf{E}_{q_\lambda(B^c|\theta,D)}[ \mathsf{K}[q_\lambda(\theta|D),t(\theta)]]
+\mathsf{E}_{q_\lambda(\theta|D)}[\mathsf{K}[q_\lambda(B^c|\theta,D),r(B^c)]] \nonumber \\
&=\overbrace{\mathsf{E}_{q(\theta,B^c|D)}[-\log(\ell(Y|f(\theta,B^c,X)))]}^\text{Expected Loss} \nonumber \\
&+ \underbrace{\mathsf{K}[q_\lambda(\theta|D),t(\theta)]}_\text{Global Regularizer}
+\underbrace{\mathsf{E}_{q_\lambda(\theta|D)}[\mathsf{K}[q_\lambda(B^c|\theta,D),r(B^c)]]}_\text{Local Regularizer} \nonumber \\
=& \mathsf{E}_{q(\theta,B^c|D)}[-\log(\ell(Y|f(\theta,B^c,X)))]+\mathsf{K}[q(\theta,B^c|D), t(\theta)r(B^c))]
\end{align}

\section{Proof of Corollary \ref{coro:PAC-Bayes}}
\label{appendix:C}

The proof of this corollary is derived from the proof of Theorem 3 in (\cite{PAC-Bayes-Germain}). More specifically, Equation \ref{subeq:Jensen-inequality} comes from Jensen inequality, Equation \ref{subeq:Donsker-Varadhan} is a result of Donsker-Varadhan change of measure inequality, and Equation \ref{subeq:markov} comes from Markov's inequality.

\begin{align}
    &\eta\mathsf{E}_\nu [-\log\big(\ell(Y|X)\big)] = \eta\mathsf{E}_\nu [-\log\big(\mathsf{E}_{q(\theta,B^c|X,Y)}[\ell(Y|X,\theta,B^c)\big)]] \nonumber \\ 
    & \leq \eta\mathsf{E}_\nu [\mathsf{E}_{q(\theta,B^c|X,Y)}[-\log\big(\ell(Y|X,\theta,B^c)\big)]] \label{subeq:Jensen-inequality}\\
    & \leq \eta\mathsf{E}_{\nu^c} [\mathsf{E}_{q(\theta,B^c|X,Y)}[-\log\big(\ell(Y|X,\theta,B^c)\big)]] \nonumber \\
    & + \mathsf{K}[q(\theta,B^c|X,Y), \pi(\theta, B^c)] \nonumber \\
   &  + \log\bigg(\mathsf{E}_{\pi(\theta,B^c)}[\exp\bigg(\eta\mathsf{E}_\nu[-\log(\ell(Y|X,\theta,B^c))]-\eta\mathsf{E}_{\nu^c}[-\log(\ell(Y|X,\theta,B^c))]\bigg)]\bigg) \label{subeq:Donsker-Varadhan} \\
   &  \stackrel{\leq}{w.p >1-\delta}\; \;  \eta\mathsf{E}_{\nu^c} [\mathsf{E}_{q(\theta,B^c|X,Y)}[-\log\big(\ell(Y|X,\theta,B^c)\big)]] + \mathsf{K}[q(\theta,B^c|X,Y), \pi(\theta, B^c)] \nonumber \\ &+\log\big(\tfrac{1}{\delta}\mathsf{E}_{\nu^c}\mathsf{E}_{\pi(\theta,B^c)}\big[\exp\bigg(\eta\mathsf{E}_\nu[-\log(\ell(Y|X,\theta,B^c))]-\eta\mathsf{E}_{\nu^c}[-\log(\ell(Y|X,\theta,B^c))]\bigg)\big]\big) \label{subeq:markov}
\end{align}

\section{Detailed Architectures of Both FEMNIST and CIFAR-100 Experiments}
\label{appendix:D}

This appendix is devoted to the mathematical representation of our proposed architecture for FedVI algorithm in Section \ref{sec:eval} along with more experimental details for both FEMNIST and CIFAR-100 experiments. For this purpose let us define:
\begin{align}
\X &\defeq \reals^{i\times i\times j} && \text{(input images; whitened)}\\
\Y&\defeq [\zeta] && \text{(labels)}\\
%f_\theta &: \reals^d \to \reals^{128} && \text{(head model; mlp with dropout)}\\
h_{\theta'}(.) &: \X \to \reals^d && \text{(embedding model; relu-convnet with dropout)}\\
g_{\theta''}(.) &: \reals^{102} \to \reals^{(2\cdot26+1)\cdot |\Y|} && \text{(posterior constructor model; relu-mlp)}\\
\theta &= \theta' \cup \theta'' \cup \theta'''  && \text{(global parameters)}\\
\beta& && \text{(local parameters)},
\end{align}
where for FEMNIST we have $i=28$, $j=1$, and $\zeta=62$, and for CIFAR-100 $i=32$, $j=3$, and $\zeta=100$, for both datasets $d=128$ and the number of local and global features are equal to $26$ and $102$, respectively\footnote{After passing the data through the embedding model, for both FEMNIST and CIFAR-100 experiments we use the first $\%80$ of the latent features as global features and the remaining $\%20$ as local features, as it is illustrated in Figure \ref{fig:divide_glob_loc_features}. More specifically, in these experiments that the dimension of the last layer of the embedding model is equal to $d=128$, the first $102$ features are considered as the global features and the rest of $26$ features are local features. }. 

%In our experiments the embedding model, $h_{\theta'}(.)$, and the posterior constructor model, $g_{\theta''}(.)$, are defined as follows:

%- $f_\theta(z)=z,$ i.e., the identity function.\\
%\begin{enumerate}
\textbf{Embedding Model:} In our experiments the embedding model, $h_{\theta'}(.)$, is a relu convnet. For FEMNIST experiment we consider the convolutional model with $2$ convolution layers that is described in Table $4$ of (\cite{FedOpt}) paper (without the top layer) and is parameteraized by the global parameters. the detailed structure of this embedding model is as the following. 

For FEMNIST: $h_{\theta'}(.)$  \textit{= conv(32) $\to$ relu $\to$ conv(64) $\to$ relu $\to$ maxpool(2,2) $\to$ dropout(0.25) $\to$ flatten $\to$ dense(128) $\to$ dropout(0.5)}

We choose a convolutional embedding model for CIFAR-100 as well, which is similar to FEMNIST embedding model, but having $5$ convolution layers instead. The detailed structure is as follows.  

For CIFAR-100: $h_{\theta'}(.)$ \textit{= conv(32) $\to$ relu $\to$ conv(64) $\to$ relu $\to$ conv(128) $\to$ relu $\to$ conv(256) $\to$ relu $\to$ conv(512) $\to$ relu $\to$ maxpool(2,2) $\to$ dropout(0.25) $\to$ flatten $\to$ dense(128) $\to$ dropout(0.5)}

\textbf{Posterior Constructor Model:} The posterior constructor model, $g_{\theta''}(.)$, is an MLP with three (dense) layers that takes the global features of the output of $h_{\theta'}(.)$ as input and generates mean, variance, and bias of the posterior.

For both FEMNIST and CIFAR-100: $g_{\theta''}(.)$  \textit{= dense(256) $\to$ relu $\to$ dense(256) $\to$ relu $\to$ dense(${(2 \times 26+1)\times |\Y|}$)}

\textbf{Global and Local Classifiers:} As it is mentioned in section \ref{sec:eval}, for both FEMNIST and CIFAR-100 experiments global classifier is one dense layer with $|\Y|$ units and no activation function, parameterized by the global parameters, and the local classifier is one dense layer similar to the global classifier, but parameterized by the local parameters. 

\textbf{Per-MiniBatch Loss Function:} Let $B_{tk}$ denote a subset of the $k^{th}$ client's evidence at $t^{th}$ epoch, where $|B_{tk}|=256$ in all of our experiments for both training and test/evaluation procedures. For CIFAR-100 training data each input image $x$ is randomly cropped to $24\times 24\times 3$ and subsequently padded to $32\times 32\times 3$. We note that there is no cropping for both training and test sets of FEMNIST data and test set of CIFAR-100.

Based on the above explanations, the per-minibatch loss is defined as:
\begin{align}
\begin{split}
\mathcal{L}_t &= \sum_{k\in[c]} \left( \tfrac{\tau}{256} \K\left[ q(\beta|\{[h_{\theta'}(x)]_{27:128}:x\in B_{tk}\}), r(\beta) \right]\right. \\
  &\left. +\sum_{(x,y) \in B_{tk}}-\log \softmax_{y}\left( \reshape(\beta,[|\Y|,26]) \cdot [h_{\theta'}(x)]^T_{1:26} \right. \right. \\
  &\left. \left. + \theta''' \cdot [h_{\theta'}(x)]^T_{27:128} + b_\beta+b_{\theta''}\right)\right)
\end{split}
\end{align}

where $\tau=10^{-9}$ for FEMNIST and $\tau=10^{-3}$ for CIFAR-100, and

\begin{align}
r(\beta) = \MVN\left(0,\sigma_0 \eye_{26\cdot |\Y|}\right)\\
q(\beta|\{[h_{\theta'}(x)]_{27:128}:x\in B_{tk}\}) &= \MVN(\mu_{tk}, \diag(\sigma_{tk}))\\
\mu_{tk} = \tfrac{1}{10} \left(\tfrac{1}{|B_{tk}|}\sum_{x\in B_{tk}} [g_{\theta''}\of h_{\theta'} \of x]_{1:(26\cdot |\Y|)}\right) & \text{(Approx equals zero at init.)}\\
\sigma_{tk} = 10^{-5}+\sigma_0 e^{ \tfrac{1}{100}\left(\frac{1}{|B_{tk}|}\sum_{x\in B_{tk}} [g_{\theta''}\of h_{\theta'}\of x]_{(1+26\cdot |\Y|):(2\cdot 26\cdot |\Y|)}\right)} & \text{(Approx equals $\sigma_0\eye_{26\cdot|\Y|}$ at init.)}\\
b_\beta = [g_{\theta''}\of h_{\theta'}\of x]_{(1+2\cdot 26\cdot |\Y|):(1+3\cdot 26\cdot|\Y|)}
\end{align}

where the $\sigma_0=\sqrt{\tfrac{2}{26+|\Y|}}$ factor exists because this is the Glorot scale \citep[e.g.][]{Glorot2010UnderstandingTD, pmlr-v15-glorot11a, He2015DelvingDI} and {$b_{\theta''}$ is the existing bias in the posterior constructor Keras model}.  

\section{Information Theoretic Perspective}
\label{appendix:E}

Neural networks from information theoretic point of view are introduced in \citep{tishby2000information}. Borrowing idea from this work, VIB \citep{alemi2017deep} reframes Variational Inference (VI) under the lens of representation learning and information theory.  In particular, their paper assumes that there exists a latent variable $z$ which is used to encode information about the input data $x$ and is used to predict the output label $y$.  They then show that training with the traditional VI objective (ELBO) amounts to maximize a lower bound on $I(y; z)$, the mutual information between the label and representation (the lower bound is given by the expected log-likelihood of the label computed over the representation).  At the same time, the KL penalty placed on the representation amounts to minimize an upper bound on $I(x; z)$, the mutual information between the input data and the latent representation. They go on to show experimentally that there is a trade-off between the information learned about the input (Rate) and the information about the output (Distortion).  Notably, their analysis further shows that the models which have a particularly high rate fail to generalize properly.  They find that the most performant models on MNIST encode roughly $10$ bits per image, while models which record less than $\log_2(10)$ bits perform extremely poorly (because they lack the info to distinguish between classes). Follow up works \citep{ThermoDynamML}, \citep{pmlr-v118-alemi20a} study different settings of this to show what happens when e.g., one uses VI to fit a BNN.  Notably, from this work, we see that $KL[q(\theta|D, r(\theta))]$ is an upper bound on the mutual information between the parameters $\theta$ and the dataset $D_i\{x_i, y_i\}$.  For our purposes, this shows that VI minimizes a bound on the information about the dataset learned by the model. 
%We rewrite the loss function of our proposed model in terms of mutual information in Section \ref{sec:infothe} to validate this intuition mathematically. \textcolor{blue}{Moreover in Section \ref{sec:eval} we compute the amounts of information learned by FedVI algorithm (which is trained by variational inference) on clients to see how much information needs to be gathered in order to offer a model which performs comparable to the baseline.}
%Thus, one potential experiment for us could be to compute the information learned by a global VI model on clients and see how much information needs to be gathered in order to offer a model which performs comparable to the baseline.
%Information harvesting is one of the important aspects of each learning model, which tells us how much information our model needs to learn to be able to make predictions. 
Therefore, using the definition of mutual information, we can rewrite the ELBO upper bound (Equation \ref{eq:loss}) in terms of mutual information as the following.
\begin{align}
\label{eq:info_theo_loss}
 \mathcal{J}(\lambda;\gamma',\tau')=\gamma' I(\theta;D) + \tau' I(B^c;\theta,D)-I(Y;\theta,B^c,X)
\end{align}
where $\gamma'$ and $\tau'$ are assumed as the hyper parameters.

 %which is a mathematical representation of what we discussed in Section \ref{sec:relatedworks-infothe}.

\textbf{proof:}

This proof is provided based on the definition of mutual information in terms of KL divergence, properties of logarithms, and multiplication rule in probability. 
\begin{align}
p(\{y^{n_k}\}^c|\{x^{n_k}\}^c)
&=\frac{p(\theta,B^c,\{y^{n_k}\}^c|\{x^{n_k}\}^c)}{p(\theta,B^c|\{y^{n_k},x^{n_k}\}^c)} %\nonumber \\
 = \frac{p(\theta,B^c,\{y^{n_k}\}^c|\{x^{n_k}\}^c)}{p(\theta,B^c|D)} \times \frac{q(\theta,B^c|D)}{q(\theta,B^c|D)} \nonumber \\
&= \frac{p(\theta,B^c,\{y^{n_k}\}^c|\{x^{n_k}\}^c)}{q(\theta,B^c|D)} \times \frac{q(\theta,B^c|D)}{p(\theta,B^c|D)} \nonumber \\
&= \frac{t(\theta)r(B^c)\ell(Y|f(\theta,B^c,X))}{q_\lambda(B^c|D)q_\lambda(\theta|B^c,D)} \times \frac{q(\theta,B^c|D)}{p(\theta,B^c|\theta,D)} \nonumber \\
&\Rightarrow -\log(p(\{y^{n_k}\}^c|\{x^{n_k}\}^c)) = -\log(\frac{q(\theta,B^c|D)}{p(\theta,B^c|c)}) \nonumber \\
&-\log(\ell(Y|f(\theta,B^c,X))+\log(\frac{q_\lambda(B^c|D)}{r(B^c)})+\log(\frac{q_\lambda(\theta|B^c,D)}{t(\theta)}) \nonumber \\
&=-\log(\frac{q(\theta,B^c|D)}{p(\theta,B^c|c)})-\log(\ell(Y|f(\theta,B^c,X)) \nonumber \\
&+\log(\frac{q_\lambda(B^c,D)}{r(B^c)p(D)})+\log(\frac{q_\lambda(\theta,B^c,D)}{t(\theta)p(B^c,D)}) \nonumber \\
&\Rightarrow \mathsf{E}_{q_\lambda(\theta,\B^c|D)}[-\log(p(\{y^{n_k}\}^c|\{x^{n_k}\}^c))+\log(\frac{q(\theta,B^c|D)}{p(\theta,B^c|c)})] \nonumber \\
&= \mathsf{E}_{q_\lambda(\theta,\B^c|D)}[-\log(\ell(Y|f(\theta,B^c,x))+\log(\frac{q_\lambda(B^c,D)}{r(B^c)p(D)})+\log(\frac{q_\lambda(\theta,B^c,D)}{t(\theta)p(B^c,D)})] \nonumber \\
&=\underbrace{\frac{1}{p(D)}}_\text{Constant}\mathsf{E}_{q_\lambda(\theta,\B^c,D)}[-\log(\frac{p(f(\theta,B^c,X)|Y)p(Y)}{p(f(\theta,B^c,x))}) \nonumber\\
&+\log(\frac{q_\lambda(B^c,D)}{r(B^c)p(D)})+\log(\frac{q_\lambda(\theta,B^c,D)}{t(\theta)p(B^c,D)})] \nonumber %\\
\end{align}
\begin{align}
&=\frac{1}{p(D)}\mathsf{E}_{q_\lambda(\theta,\B^c,D)}[-\log(\frac{p(Y,f(\theta,B^c,X))}{p(f(\theta,B^c,x))p(Y)})-\log(p(Y))] \nonumber \\
&+\frac{1}{p(D)}\mathsf{E}_{q(\theta,B^c,D)}[\log(\frac{q_\lambda(B^c,D)}{r(B^c)p(D)})]+\frac{1}{p(D)}\mathsf{E}_{q(\theta,B^c,D)}[\log(\frac{q_\lambda(\theta,B^c,D)}{t(\theta)p(B^c,D)})] \nonumber \\
&=\frac{1}{p(D)}\mathsf{E}_{q_\lambda(\theta,\B^c,D)}[-\log(\frac{p(Y,\theta,B^c,X)}{p(\theta,B^c,X)p(Y)})-\log(p(Y)] \nonumber \\
&+\frac{1}{p(D)}\mathsf{E}_{q(B^c,D)}[\log(\frac{q_\lambda(B^c,D)}{r(B^c)p(D)})]+\frac{1}{p(D)}\mathsf{E}_{q(\theta,B^c,D)}[\log(\frac{q_\lambda(\theta,B^c,D)}{t(\theta)p(B^c,D)})] \nonumber \\
&=\frac{1}{p(D)}(I(B^c;D) + I(\theta;B^c,D)-I(Y;\theta,B^c,X))-\log(p(Y)),
\end{align}

where $\frac{1}{p(D)}$ is a constant and $-\log(p(Y))$ is independent of our optimization problem so can be ignored. Therefore, we can have the following objective function 
\begin{align}\label{eq:mutual-info-loss}
     \mathcal{J}(\lambda;\gamma',\eta')&=\gamma'I(B^c;D) + \tau' I(\theta;B^c,D)-I(Y;\theta,B^c,X) \nonumber \\
     &=\alpha'I(\theta,B^c;D)-I(Y;\theta,B^c,X),
\end{align}
where, $\gamma'$, $\tau'$ and $\alpha'$ are the hyper parameters.
\hfill $\Box$
%Minimizing this objective function is equivalent to maximizing the mutual information between the labels and the joint of data, local and global parameters, minimizing the mutual information between local parameters and data, and minimizing mutual information between global parameters and the joint of local parameters and data, simultaneously. 

As we can see, by minimizing this loss function we are minimizing the mutual information between model parameters and training data, and maximizing the mutual information between labels and the joint of model parameters and training data which resembles the results of \citep{tishby2000information} in a federated setup.

In future work this representation of the loss function can be used to study FedVI algorithm from information harvesting point of view to determine how much information our algorithm needs to be able to make predictions. Moreover, considering the definition of  $\epsilon$-mutual information differential privacy in Definition 2 of \citep{Cuff_2016}, we can investigate whether minimizing this loss function (Equation \ref{eq:mutual-info-loss}) guarantees any level of differential privacy. And if not, how we should modify the loss function for data privacy purposes.  
\end{appendices}

\end{document}